\documentclass[runningheads]{llncs}
\usepackage{graphicx}
\usepackage{savesym}
\pdfoutput=1
\usepackage{amsmath,amssymb} 
\usepackage{color}
\usepackage[width=122mm,left=12mm,paperwidth=146mm,height=193mm,top=12mm,paperheight=217mm]{geometry}

\usepackage{amsfonts}
\usepackage{algorithm}
\usepackage{algorithmic}
\usepackage{enumitem}
\usepackage{subfig}
\usepackage{widetext}
\usepackage{bbm}
\usepackage{url}
\usepackage{mysymbols}
\usepackage{multirow}
\usepackage{array}
\usepackage{hyperref}
\newcolumntype{M}{>{\centering\arraybackslash}m{\dimexpr.25\linewidth-2\tabcolsep}}
\begin{document}
\pagestyle{headings}
\mainmatter

\title{GeThR-Net: A Generalized Temporally Hybrid Recurrent Neural Network for Multimodal Information Fusion} 

\titlerunning{GeThR-Net}

\authorrunning{Ankit Gandhi \textit{et al.}}

\author{Ankit Gandhi$^{1\,*}$, Arjun Sharma$^{1\,*}$
, Arijit Biswas$^2$, \and Om  Deshmukh$^1$}
\institute{$^1$ Xerox Research Centre India; $^2$ Amazon Development Center India \\
\email{\{ankit.g1290,arjunsharma.iitg,arijitbiswas87\}@gmail.com; om.deshmukh@xerox.com} (*-equal contribution)}

\maketitle

\begin{abstract}
Data generated from real world events are usually temporal and contain multimodal information such as audio, visual, depth, sensor etc. which are required to be intelligently combined for classification tasks. In this paper, we propose a novel generalized deep neural network architecture where temporal streams from multiple modalities are combined. There are total M+1 (M is the number of modalities) components in the proposed network. The first component is a novel temporally hybrid Recurrent Neural Network (RNN) that exploits the complimentary nature of the multimodal temporal information by allowing the network to learn both modality specific temporal dynamics as well as the dynamics in a multimodal feature space. M additional components are added to the network which extract discriminative but non-temporal cues from each modality. Finally, the predictions from all of these components are linearly combined using a set of automatically learned weights. We perform exhaustive experiments on three different datasets spanning four modalities. The proposed network is relatively 3.5\%, 5.7\% and 2\% better than the best performing temporal multimodal baseline for UCF-101, CCV and Multimodal Gesture datasets respectively. 

\end{abstract}

\section{Introduction}

Humans typically perceive the world through multimodal sensory information~\cite{stein2009neural} such as visual, audio, depth, etc.. For example, when a person is running, we recognize the event by looking at how the body posture of the person is changing with time as well by listening to the periodic sound of his/her footsteps.  Human brains can seamlessly process multimodal signals and accurately classify an event or an action. However, it is a challenging task for machines to exploit the complimentary nature and optimally combine multimodal information.


Recently, deep neural networks have been extensively used in computer vision, natural language processing and speech processing. LSTM~\cite{hochreiter1997long}, a Recurrent Neural Network (RNN)~\cite{williams1989learning} architecture, has been extremely successful in temporal modelling and classification tasks such as handwriting recognition~\cite{graves2009offline}, action recognition~\cite{baccouche2011sequential}, image and video captioning \cite{Chen_2015_CVPR,Vinyals_2015_CVPR,yao2015describing} and speech recognition~\cite{graves2013speech,graves2014towards}. RNNs can also be used to model multimodal information. These methods fall under two broad categories:  (a) Early-Fusion: modality specific features are combined to create a feature representation and fed into a LSTM network for classification. (b) Late-Fusion: each modality is modelled using individual LSTM networks and their predictions are combined for classification \cite{wu2015modeling}. Since early-fusion techniques do not learn any modality specific temporal dynamics, they fail to capture the discriminative temporal cues present in each modality. On the other hand, late-fusion methods cannot extract the discriminative temporal cues which might be available in a multimodal feature representation. In this paper, we propose a novel generalized temporally hybrid Recurrent Neural Network architecture called GeThR-Net which models the temporal dynamics of individual modalities (late fusion) as well as the overall temporal dynamics in a multimodal feature space (early fusion). 

GeThR-Net has one temporal and $M$ ($M$ is the total number of modalities) non-temporal components. The novel temporal component of GeThR-Net models the long-term temporal information in a multimodal signal whereas the non-temporal components take care of situations where explicit temporal modelling is difficult. The temporal component consists of three layers. The first layer models each modality using individual modality-specific LSTM networks. The second layer combines the hidden representations from these LSTMs to form a multimodal feature representations corresponding to each time step. In the final layer, one multimodal LSTM is trained on the multimodal features obtained from the second layer. The output from the final layer is fed into a softmax layer for category-wise confidence prediction. We observe that in many real world scenarios, the temporal modelling of individual or multimodal information is extremely hard due to the presence of noise or high intra-class temporal variation. We address this issue by introducing additional $M$ components to GeThR-Net which model modality specific non-temporal cues by ignoring the temporal relationship across features extracted from different time-instants. The predictions corresponding to all $M+1$ components in the proposed network are combined using a weighted vector learned from the validation dataset. We note that GeThR-Net can be used with any kind of modality information without any restriction on the number of modalities. 

The main contributions of this paper are:
\noindent
\begin{itemize}[noitemsep,leftmargin=*]
\item We propose a generalized deep neural network architecture called GeThR-Net that could intelligently combine multimodal temporal information from any kind and from any number of streams.
\item Our objective is to propose a general framework that could work with modalities of any kind. We demonstrate the effectiveness and wide applicability of GeThR-Net by evaluation of classification performance on three different action and gesture classification tasks, UCF-101~\cite{soomro2012ucf101}, Multimodal Gesture~\cite{escalera2013multi} and Columbia Consumer videos~\cite{jiang2011consumer}. Four different modalities such as audio, appearance, short-term motion and skeleton are considered in our experiments. We find out that GeThR-Net is relatively 3.5\%, 5.7\% and 2\% better than the best temporal multimodal baseline for UCF-101, CCV and Multimodal Gesture datasets respectively.
\end{itemize}

\begin{figure*}[t]
\centering
\includegraphics[width=0.8\linewidth]{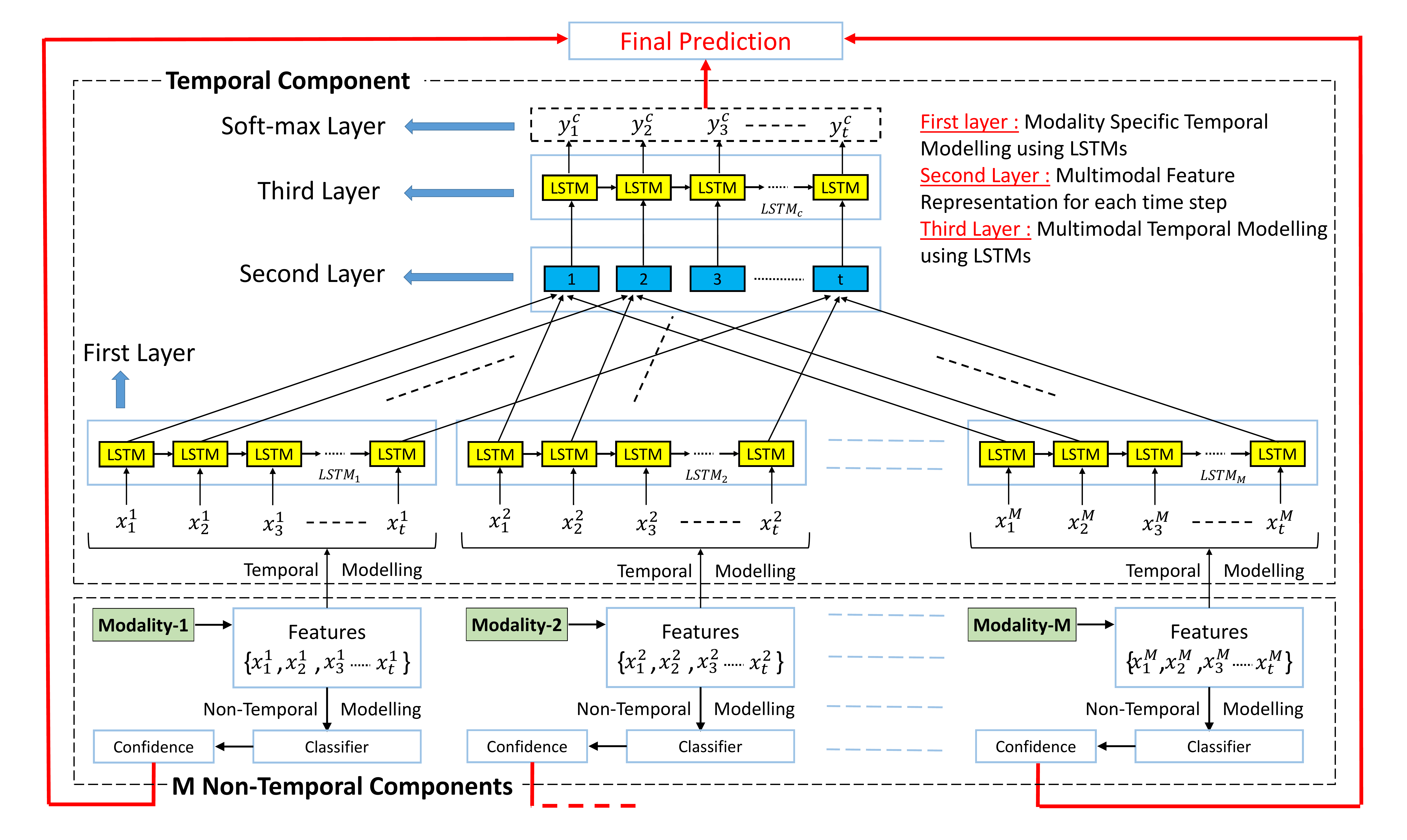}
\caption{The overall pipeline of the proposed approach GeThR-Net. The input to the system is a multimodal stream (e.g.: appearance, short-term motion, skeleton and/or audio for action/gesture classification tasks) and output is the class label. The proposed network has total $M+1$ components ($M$ is the total number of modalities). The first component is a temporally hybrid network that models the modality specific temporal dynamics as well as the temporal dynamics in a multimodal feature space. Corresponding to each of the $M$ modalities, there is also a non-temporal classification component in the network. All of these components in the network are trained in an end-to-end fashion.}
\label{fig:pipeline}
\end{figure*}

The full pipeline of the proposed approach is shown in Figure \ref{fig:pipeline}. We discuss the relevant prior work in Section \ref{sec:related_work} followed by the details of GeThR-Net in Section \ref{sec:proposed_approach}. The details of experimental results are provided in Section \ref{sec:exp_results}.

\section{Related Work}
\label{sec:related_work}
In this section, we describe the relevant prior work on generic multimodal fusion and multimodal fusion using deep learning.

\underline{\bf Multimodal Information Fusion:} A good survey of different fusion strategies for multimodal information is in \cite{atrey2010multimodal}. We discuss a few relevant papers here. The authors in \cite{xie2013multimodal} provide a general theoretical analysis for multimodal information fusion and implements novel information theoretic tools for multimedia applications. \cite{wu2004optimal} proposes a two-step approach for an optimal multimodal fusion, where in the first step statistically independent modalities are found from raw features and in the second step, super-kernel fusion is used to find the optimal combination of individual modalities. In \cite{jhuo2014discovering}, the authors propose a method for detecting complex events in videos by using a new representation, called bi-modal words, to explore the representative joint audio and visual patterns. \cite{jiang2009short} proposes a method to extract a novel representation, the Short-term Audio-Visual Atom (S-AVA), for improved semantic concept detection in videos. The authors in \cite{ye2012robust} propose a rank minimization method to fuse the predicted confidence scores of multiple models based on different kinds of features. Their goal is to find a shared rank-2 pairwise relationship matrix (for the test samples) based on which each original score matrix from individual model can be decomposed into the common rank-2 matrix and sparse deviation errors. \cite{snoek2005early} proposes an early and a late fusion scheme for audio, visual and textual information  fusion for semantic video analysis and demonstrates that the late fusion method works slightly better. In  \cite{paleari2006toward}, the authors propose a multimodal fusion technique and describe a way to implement a generic framework for multimodal emotion recognition.

\underline{\bf Deep Learning for Multimodal Fusion:} In \cite{ngiam2011multimodal}, the authors propose a deep autoencoder network that is pretrained using sparse Restricted Boltzmann Machines (RBM). The proposed method is used to learn multimodal feature representation for the task of audio-visual speech recognition.
The authors in \cite{srivastava2012multimodal}, propose a Deep Boltzmann Machine (DBM)  for learning a generative model of data that consists of multiple and diverse input modalities. \cite{sohn2014improved}, proposes a multimodal representation learning framework that minimizes the variation information between data modalities through shared latent representations. In \cite{wu2014exploring}, the authors propose a unified  deep neural network, which jointly learns feature relationships and class relationships, and simultaneously carries out video classification within the same framework utilizing the learned relationships. \cite{mao2014explain,mao2014deep} proposes an approach for generating novel image captions given an image. This approach directly models the probability distribution of a word given previous words and an image using a network that consists of a deep RNN for sentences and a deep CNN for images. \cite{wu2014multimodal} proposes a novel bi-modal dynamic network for gesture recognition. High level audio and skeletal joints representations, extracted using dynamic Deep Belief Networks (DBN), are combined using a layer of perceptron. However, none of these approaches use RNNs for both multimodal and temporal data fusion and hence cannot learn features which truly represent the complimentary nature of multimodal features along the temporal dimension. The authors in \cite{chen2015multi}, propose a multi-layer RNN for multi-modal emotion recognition. However, the number of layers in the proposed architecture is equal to the number of modalities, which restricts the maximum number of modalities which can be used simultaneously. The authors in \cite{wu2015modeling} propose a hybrid deep learning framework for video classification that can model static spatial information, short-term motion, as well as long-term temporal clues in the videos. The spatial and the short-term motion features extracted from CNNs are combined using a regularized feature fusion network. LSTM is used to model only the modality specific long-term temporal information. However, in the proposed GeThR-Net, the temporally hybrid architecture can automatically combine temporal information from multiple modalities without requiring any explicit feature fusion framework.  We also point out that unlike \cite{wu2015modeling}, in GeThR-Net, the multimodal fusion is performed at the LSTM network level. 

To the best of authors' knowledge, there are no prior approaches where multimodal information fusion is performed at the RNN/LSTM level. GeThR-Net is the first method to use a temporally hybrid RNN which is capable of learning features from modalities of any kind without any upper-bound on the number of modalities.  

\section{Proposed Approach}
\label{sec:proposed_approach}
In this section, we provide the details of the proposed deep neural network architecture GeThR-Net. First, we discuss how LSTM networks usually work. Next, we provide the descriptions of the temporal and non-temporal components of our network followed by how we combine predictions from all these components. 

\subsection{Long Short Term Memory Networks}
\label{sec:lstm}

Recently, a type of RNN, called Long Short Term Memory (LSTM) Networks, have been successfully employed to capture long-term temporal patterns and dependencies in videos for tasks such as video description generation, activity recognition etc. RNNs~\cite{williams1989learning} are a special class of artificial neural networks, where cyclic connections are also allowed. These connections allow the networks to maintain a memory of the previous inputs, making them suitable for modelling sequential data. In LSTMs, this memory is maintained with the help of three non-linear multiplicative gates which control the in-flow, out-flow, and accumulation of information over time.  We provide a detailed description of RNNs and LSTM networks below.

Given an input sequence $\textbf{x}=\{x_{t}\}$ of length $T$, the fixed length hidden state or memory of an RNN \textbf{h} is given by
\begin{equation}
h_{t} = g(x_{t}, h_{t-1}) \quad t = 1,\dots,T 
\label{eq:hidden}
\end{equation}
We use $h_{0} = 0$ in this work. Multiple such hidden layers can be stacked on top of each other, with $x_t$ in equation~\ref{eq:hidden} replaced with the activation at time $t$ of the previous hidden layer, to obtain a `deep' recurrent neural network. The output of the RNN at time $t$ is computed using the state of the last hidden layer at $t$ as
\begin{equation}
y_{t} = \theta(W_{yh}h_{t}^{n} + b_{y})
\label{eq:output}
\end{equation}
where $\theta$ is a non-linear operation such as sigmoid or hyperbolic tangent for binary classification or softmax for multiclass classification, $b_y$ is the bias term for the output layer and $n$ is the number of hidden layers in the architecture. The output of the RNN at desired time steps can then be used to compute the error and the network weights are updated based on the gradients computed using Back-propagation Through Time (BPTT).
In simple RNNs, the function $g$ is computed as a linear transformation of the input and previous hidden state, followed by an element wise non-linearity.
\begin{equation}
g(x_{t}, h_{t-1}) = \theta(W_{hx}x_{t} + W_{hh}h_{t-1} + b_{h})
\end{equation} 
Such simple RNNs, however, suffer from the vanishing and exploding gradient problem \cite{hochreiter1997long}. To address this issue, a novel form of recurrent neural networks called the Long Short Term Memory (LSTM) networks were introduced in \cite{hochreiter1997long}. The key difference between simple RNNs and LSTMs is in the computation of $g$, which is done in the latter using a memory block. An LSTM memory block consists of a memory cell $c$ and three multiplicative gates which regulate the state of the cell - forget gate $f$, input gate $i$ and output gate $o$. The memory cell encodes the knowledge of the inputs that have been observed up to that time step. The forget gate controls whether the old information should be retained or forgotten. The input gate regulates whether new information should be added to the cell state while the output gate controls which parts of the new cell state to output. 
Like simple RNNs, LSTM networks can be made deep by stacking memory blocks. The output layer of the LSTM network can then be computed using equation~\ref{eq:output}. We refer the reader to \cite{hochreiter1997long} for more technical details on LSTMs. 

\subsection{Temporal Component of GeThR-Net}
In this subsection, we describe the details of the temporal component, which is a temporally hybrid LSTM network that models modality specific temporal dynamics as well as the multimodal temporal dynamics. This network has three layers. The first layer models the modality specific temporal information using individual LSTM layers. Multimodal information do not interact with each other in this layer. In the second layer, the hidden representations from all the modalities are combined using a linear function, followed by sigmoid non-linearity, to create a single multimodal feature representation corresponding to each time step. Finally, in the third layer, a LSTM network is fed with the learned multimodal features from the second layer. The output from the third layer is fed into a softmax layer for estimating the classification confidence scores corresponding to each label. This component is fully trained in an end-to-end manner and does not require any explicit feature fusion modelling.

Now, we describe the technical details of these layers. We assume that there are total $M$ different modalities and total $T$ time-steps. The feature representation for modality $m$ corresponding to time instant $t$ is given by: $x^m_{t}$. Now, we describe the mathematical details:
\begin{itemize}[noitemsep,leftmargin=*]

\item {\bf First Layer:} The input to this layer is $x^m_{t}$ for modality $m$ at time instant $t$. If $LSTM_{m}$ denotes the LSTM layer for modality $m$ and if $h^m_{t}$ denotes the corresponding hidden representation at time $t$, then:
\begin{displaymath} h^m_{t} = LSTM_{m}(x^m_{t}) \end{displaymath}
\item {\bf Second Layer:} In this layer, the hidden representations are combined using a linear function followed by a sigmoid non-linearity. The objective of using this layer is to combine features from multiple temporal modalities. Let us assume that $z_{t}$ denotes the concatenated hidden representation from all the modalities at time-step $t$. $W_z$ (same for all time-step $t$) denotes the weight matrix which combines the multimodal features and creates a representation $p_t$ at time instant $t$. $b_z$ denotes a linear bias and $\sigma$ is the sigmoid function.
\begin{displaymath} z_{t} = (h^1_{t},\cdots,h^m_{t}), \hspace{1cm} p_{t} = \sigma(W_{z}z_{t} + b_{z}) \end{displaymath}
\item {\bf Third Layer:} In this layer, one modality-independent LSTM layer is used to model the overall temporal dynamics of the multimodal feature representation $p_{t}$. Suppose, $LSTM_{c}$ denotes the combined LSTM and $h^c_t$ denotes the hidden representation from this LSTM layer at time $t$. $W_o$ is the weight matrix that linearly transforms the hidden representation. The output is propagated through a softmax function $\theta$ to obtain the final classification confidence values $y^c_t$ at time $t$. $b_{o}$ is a linear bias vector.
\begin{displaymath} h^c_t = LSTM_{c}(p_{t}), \hspace{1cm} y^c_t = \theta(W_{o}h^c_t + b_{o}) \end{displaymath}
\end{itemize}


%

%

\subsection{Non-temporal Component of GeThR-Net}
Although it is important to model the temporal information in multimodal signals for accurate classification or any other tasks, often in real world scenarios multimodal information contains significant amount of noise and large intra-class variation along the temporal dimension. For example, videos of the activity `cooking' often contain action segments such as `changing thermostat' or `drinking water' which are no way related to the actual label of the video. In those cases, modelling only the long-term temporal information in the video could lead to inaccurate results. Hence, it is important that we allow the proposed deep network to learn the non-temporal features too. We analyze videos from multiple datasets and observe that a simple classifier which is trained on `frame-level' features (definition of frame could vary according to the features) could give a reasonable accuracy, especially when videos contain unrelated temporal segments. Please refer to Section \ref{sec:discussion_results} for more experimental results on this. Since our objective is to propose a generic deep network that could work with any kind of multimodal information, we add additional components to the GeThR-Net, which explicitly model the modality specific non-temporal information.  

During training, for each modality $m$, we train a classifier where the set $\{x^m_{t}\}$, $\forall t$ is used as the training examples corresponding to the class of the multimodal signal. While testing for a given sequence, the predictions across all the time-steps are averaged to obtain the classifier confidence scores corresponding to all of the classes. In this paper, we have explored four different modalities: appearance, short-term motion, audio (spectrogram and MFCC) and skeleton. For appearance, short-term motion and audio-spectrogram, we use fine-tuned CNNs and for audio-MFCC and skeleton, we use SVMs as the non-temporal classifiers. 
\subsection{Combination}
There are total $M+1$ components in GeThR-Net, where the first one is the temporally hybrid LSTM network and the rest $M$ are the non-temporal modality specific classifiers corresponding to each modality. Once we independently train these $M+1$ classifiers, their prediction scores are combined and a single class-label for each multimodal temporal sequence is predicted. We use a validation dataset to determine the relevant weights corresponding to each of the $M+1$ components.    

\section{Experiments}
\label{sec:exp_results}
Our goal is to demonstrate that the proposed GeThR-Net can be effectively applied to any kind of multimodal fusion. To achieve that, we perform thorough experimental evaluation and provide the details of the experimental results in this section. 

\subsection{Dataset Details}
The dataset details are provided in this subsection.

\textbf{UCF-101~\cite{soomro2012ucf101}:} UCF-101 is an action recognition dataset containing realistic action videos from YouTube. The dataset has 13,320 videos annotated into 101 different action classes. The average length of the video in this dataset is 6-7 sec. 
The dataset possess various challenges and diversity in terms of large variations in camera motion, object appearance and pose, cluttered background, illumination, viewpoint, etc. We evaluate the performance on this dataset following the standard protocol~\cite{wu2015modeling,soomro2012ucf101} by reporting the mean classification accuracy across three training and testing splits. We use the appearance and short-term motion modality for this dataset~\cite{SimonyanZ14,wu2015modeling}.

\textbf{CCV~\cite{jiang2011consumer}:} The Columbia Consumer Videos (CCV) has 9,317 YouTube videos distributed over 20 different semantic categories. The dataset has events like `baseball', `parade', `birthday', `wedding ceremony', scenes like `beach', `playground', etc. and objects like `cat', `dog' etc...  The average length of the video in this dataset is 80 sec long. For our experiments, we have used 7751 videos (3851 for training and 3900 for testing) as the remaining videos are not available on YouTube presently.  In this dataset, the performance is measured by average precision (AP) for each class and the overall measure is given by mAP (mean average precision over 20 categories). In this dataset, we use three different modalities, i.e., appearance, short-term motion and audio.

\textbf{Multimodal Gesture Dataset~\cite{escalera2013multi} (MMG):} ChaLearn-2013 multimodal gesture recognition dataset is a large video database of 13,858 gestures from a lexicon of 20 Italian gesture categories. The focus of the dataset is on user independent multiple gesture learning. The dataset has RGB and depth images of the videos, user masks, skeletal model, and the audio information (utterance of the corresponding gesture by the actor), which are synchronous with the gestures performed.  The dataset has 393 training, 287 testing, and 276 testing sequences. Each sequence is of duration between 1-2 minutes and contains 8-20 gestures. Furthermore, the test sequences also have `distracter' (out of vocabulary) gestures apart from the 20 main gesture categories. 
For this dataset, we use the audio and skeleton modality for fusion because some of the top-performing methods~\cite{escalera2013multi} on this dataset also used these two modalities. The loose temporal boundaries of the gestures in the sequence is available during training and validation phase, however, at the time of testing, the goal is to also predict the correct order of gestures within the sequence along with the gesture labels. The final evaluation is defined in terms of edit distance (insertion, deletion, or substitution) between  the ground truth sequence of labels and the predicted sequence of labels. The overall score is the sum of edit distance for all testing videos, divided by the total number of gestures in all the testing videos~\cite{escalera2013multi}.  
\subsection{Modality Specific Feature Extraction}
\label{modality_specific_feature}
In this section, we describe the feature extraction method for different modalities - appearance, short-term motion, audio, and skeleton, which are used in this paper across three different datasets.
\begin{itemize}[noitemsep,leftmargin=*]

\item{\textbf{Appearance Features:}}
We adopted the VGG-16~\cite{simonyan2014very} architecture to extract the appearance features. In this architecture, we change the number of neurons in fc7 layer from 4096 to 1024 to get a compressed lower dimensional representation of an input. We finetune the final three fully connected layers (fc6, fc7, and fc8) of the network pretrained on ImageNet using the frames of the training videos. The activations of the fc7 layer are taken as the visual representation of the frame provided as an input. While finetuning, we use    minibatch stochastic descent with a fixed momentum of 0.9. The input size of the frame to our model is 224 $\times$ 224 $\times$ 3. Simple data augmentations are also done such as cropping and mirroring~\cite{jia2014caffe}. We adopt a dropout ratio of 0.5. The initial learning rate is set to 0.001 for fc6, and 0.01 for fc7 and fc8 layers as the weights of last two layers are learned from scratch. The learning rate is reduced by factor of 10 after every 10,000 iterations. 

\item{\textbf{Short-Term Motion Features:}}
To extract the features, we adopted the method proposed in the recent two-stream CNN paper~\cite{SimonyanZ14}. This method stacks the optical flows computed between pairs of adjacent frames over a time window and provides it as an input to CNN. We used the same VGG-16 architecture (as above) with 1024 neurons in fc7 layer, and pre-training on ImageNet for the extraction of short-term motion features. However, unlike the previous case (where input to the model was an RGB image comprising of three channels), the input to this network is a 10-frame stacking of optical flow fields (x and y direction), and thus the convolution filters in the first layer are different from those of the appearance network. We adopt a high dropout rate of 0.8 and set the initial learning rate to 0.001 for all the layers. The learning rate is reduced by a factor of 10 after every 10,000 iterations.

\item{\textbf{Audio Features:}}
We use two different kinds of feature extraction method for audio modality.

\begin{itemize}[noitemsep,leftmargin=*]

\item{\bf Spectrogram Features:} In this method, we extract the spectrogram features from audio signal using a convolutional neural network~\cite{van2013deep}. We divide the video into multiple overlapping 1 sec clips and then, apply the Short Time Fourier Transformation to convert each one second 1-d audio
signal into a 2-D image (namely log-compressed mel-spectrograms
with 128 components) with the horizontal axis and vertical axis being
time-scale and frequency-scale respectively. The features
are extracted from these spectrogram images by providing them as input to a CNN. In this case, we use AlexNet~\cite{krizhevsky2012imagenet} architecture and the network was pre-trained on ImageNet. We finetune the final three layers of network with respect to the spectrogram images of training videos to
learn the `spectrogram-discriminative' CNN features. We also change the number of
nodes in fc7 layer to 1024 and use the activations of fc7 layer as the representation of a spectrogram image. The learning rate and dropout parameters are same as mentioned in the appearance feature extraction case.

\item{\bf MFCC Features:} We use MFCC features for the MMG dataset. The spectrogram based CNN features were not used for this dataset as the temporal extent of each gesture was very less (1-2 sec), making it difficult to extract multiple spectrograms along the temporal dimension. In this method, speech signal of a gesture was analyzed using a 20ms Hamming window with a fixed frame rate of 10ms. Our feature consists of 12 Mel Frequency Cepstral Coefficients (MFCCs) along with the log energy ($MFCC_0$) and their first and second order delta values to capture the spectral variation.
We concatenated 5 adjacent frames together in order to adhere to the 20fps of videos in the MMG dataset. Hence, we have a feature of dimension of $39 \times 5 = 195$ for each frame of the video. The data was also normalized such that each of the features (coefficients, energy and derivatives) extracted have zero mean and one variance.
 
\end{itemize}

\item{\textbf{Skeleton Features:}}

We use the skeleton features for the MMG dataset. We employ the feature extraction method proposed in ~\cite{wu2014multimodal,yang2012eigenjoints} to characterize the action information which includes the posture feature, motion feature and offset feature. Out of 20 skeleton joint locations, we use only  9 upper body joints as they are the most discriminative for recognizing gestures. 


\end{itemize}
\subsection{Methods Compared}
\label{sec:methods_compared}
\begin{figure*}[t]
\centering
\includegraphics[width=0.6\linewidth]{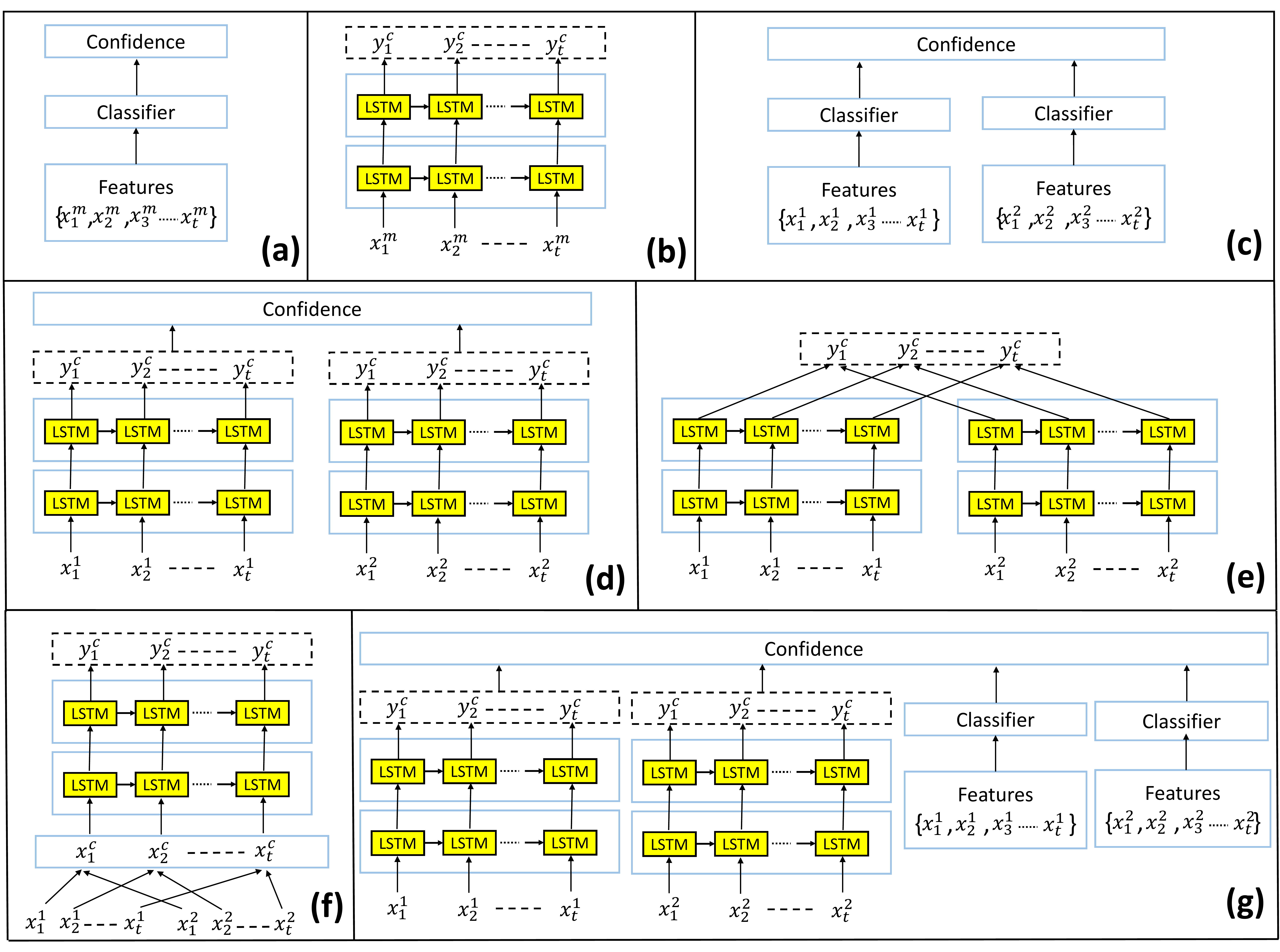}
\caption{Different baselines which are compared with GeThR-Net. (a) NonTemporal-M. (b) Temporal-M. (c) NonTemporal-AM (d) Temporal-AM (late fusion) (e) TemporalEtoE-AM (late fusion) (f) Temporal-AM (early fusion) (g)  Temporal-AM+NonTemporal-AM.}
\label{fig:figure_baseline}
\end{figure*}
To establish the efficacy of the proposed approach, we compare GeThR-Net with several baselines. The baselines were carefully designed to cover several temporal and non-temporal feature fusion methods. We provide the architectural details of these baselines in Figure \ref{fig:figure_baseline} for easy understanding of their differences. 
\begin{enumerate}[label=(\alph*),noitemsep,leftmargin=*]
\item {\bf NonTemporal-M:} In this baseline, we train modality specific non-temporal models and predict label of a temporal sequence based on the average over all predictions across time. For appearance, short-term motion and audio spectrogram, we use CNN features (Section \ref{modality_specific_feature}) followed by a softmax layer for classification. For audio MFCC and Skeleton, we use the features extracted  using the methods described in \ref{modality_specific_feature} followed by SVM classification. Multimodal fusion is not performed for label prediction in these baselines.
\item {\bf Temporal-M:} For this baseline, we feed the modality specific features (as described in the last subsection), to LSTM networks for the temporal modelling and label prediction. Here also, features from multiple modalities are not fused for classification. 
\item {\bf NonTemporal-AM (all modality combined):} In this baseline, the outputs from the modality specific non-temporal baselines (CNN/SVM) are linearly combined for classification. The combination weights are automatically learned from validation datasets. 
\item {\bf Temporal-AM (late fusion, all modality combined):} Here also, the outputs from the modality specific temporal baselines (LSTMs) are linearly combined for classification. This is a late fusion approach.
\item {\bf TemporalEtoE-AM (late fusion, all modality combined):} In this baseline, we add a linear layer on top of the modality specific temporal baselines and use an end-to-end training approach for learning the weights of the combination layer. This is also a late fusion approach.
\item {\bf Temporal-AM (early fusion, all modality combined):} Features from multiple modalities are linearly combined and then forward propagated through a LSTM for classification. This is an early fusion approach. 
\item {\bf Temporal-AM+NonTemporal-AM (all modality combined):} In this baseline, the outputs from all the modality specific temporal and nontemporal baselines are combined for the final label prediction. Here also, we use a validation dataset for predicting the optimal weights corresponding to each of these components.
\item {\bf TemporallyHybrid-AM (proposed, all modality combined):} This method uses only the temporally hybrid component of the proposed approach. The non-temporal components' outputs are not used. This network is completely trained in an end-to-end fashion (See the temporal component in Figure \ref{fig:pipeline}).
\item {\bf GeThR-Net:} This is the proposed approach (See Figure \ref{fig:pipeline}). 
\end{enumerate}
\subsection{Implementation Details}
We used the initial learning rate of 0.0002 for all LSTM networks. It is reduced by a factor of 0.9 for every epoch starting from the 6-th epoch. We set the dropout rate at 0.3. For the baseline methods of temporal modelling, Temporal-M, Temporal-AM and TemporalEtoE-AM, we tried different combinations for the number of hidden layers and the number of units in each layer and chose the one which led to the optimal performance on the validation set. Since, the feature dimension is high (1024) in UCF-101 and CCV dataset, the number of units in each layer is varied from 256 to 768 in the intervals of 32. While in case of MMG, it is varied from 64 to 512 in the same interval. The number of layers in the baselines were varied between 1 and 3 for all of the datasets.

For the proposed temporally hybrid network (TemporallyHybrid-AM) component also, the number of units in the First-layer LSTM corresponding to each modality, the number units in the linear Second-layer and the number of units in Third-layer multimodal LSTM are chosen based upon the performance on the validation dataset. For UCF-101 dataset, the First-layer has 576 units for both the appearance and short-term modality. The Second-layer has 768 units and the Third-layer has 448 units. For CCV dataset, all the three modalities, appearance, short-term motion and audio have 512 units in the First-layer. In CCV, the Second-layer has 896 units and the Third-layer has 640 units. For MMG dataset, the First-layer has 256 units for skeleton modality and 192 units for audio modality. The Second-layer has 384 units and the Third-layer has 256 units. Note that these parameters differ across the datasets due to the variation in the input feature size and the inherent complexity of the datasets. 

\begin{table}[t]
\begin{center}
\caption{Comparison of GeThR-Net with baseline methods on UCF-101, CCV and Multimodal Gesture recognition (MMG) dataset. UCF-101: M1 is appearance, M2 is short-term motion and classification accuracy is reported. CCV: M1 is appearance, M2 is short-term motion, M3 is audio and mean average precision (mAP) is reported. MMG: M1 is audio, M2 is skeleton and normalized edit distance is reported.}

\begin{tabular}{|c|c|}
\hline
Dataset & Modalities Used \\ \hline \hline
 & Appearance \\
\multirow{1}{*}{UCF} & (M1) \\ \cline{2-2}
\multirow{1}{*}{-101} & Short term Motion \\
& (M2)  \\ \hline
\multirow{6}{*}{CCV} & Appearance \\
& (M1)  \\ \cline{2-2}
& Short-term Motion \\
& (M2)  \\ \cline{2-2}
& Audio \\
& (M3)  \\ \hline
\multirow{4}{*}{MMG} & Audio \\
& (M1)  \\ \cline{2-2}
& Skeleton \\
& (M2)  \\ \hline
\end{tabular}
\quad
\begin{tabular}{|l|c|c|c|}
\hline
Methods & \multirow{1}{*}{UCF-101} & \multirow{1}{*}{CCV} & \multirow{1}{*}{MMG}\\
& \multirow{1}{*}{(Accuracy)} & \multirow{1}{*}{(mAP)} & \multirow{1}{*}{(edit)} \\ 	\hline \hline
NonTemporal-M1 & 76.3 & 76.7 & 0.988  \\ \hline
NonTemporal-M2 & 86.8 & 57.3 & 0.782 \\ \hline
NonTemporal-M3 & -  & 30.3 & - \\ \hline \hline
Temporal-M1 & 76.6 & 71.7 & 0.284 \\ \hline
Temporal-M2 & 85.5 & 55.1 & 0.361 \\ \hline
Temporal-M3 & - & 28.5 & - \\ \hline \hline
NonTemporal-AM & 89.9 & 78.5 & 0.776 \\ \hline \hline
\multirow{1}{*}{Temporal-AM} & \multirow{2}{*}{88.0} & \multirow{2}{*}{75.0}  & \multirow{2}{*}{0.156} \\
\multirow{1}{*}{(late fusion)} &  &  & \\ \hline
TemporalEtoE-AM & \multirow{2}{*}{88.4} & \multirow{2}{*}{72.5} & \multirow{2}{*}{0.155} \\  
\multirow{1}{*}{(late fusion)} &  &  & \\ \hline 
\multirow{1}{*}{Temporal-AM} & \multirow{2}{*}{86.5}	& \multirow{2}{*}{73.1} & \multirow{2}{*}{0.190} \\
\multirow{1}{*}{(early fusion)} &   &  & \\ \hline \hline
\multirow{1}{*}{Temporal-AM +} & \multirow{2}{*}{90.2} & \multirow{2}{*}{79.2} & \multirow{2}{*}{0.155} \\
\multirow{1}{*}{NonTemporal-AM} &  &  & \\ \hline \hline
TemporallyHybrid-AM & 89.0 & 74.0 & {\bf 0.152} \\ \hline
GeThR-Net & {\bf 91.1} & {\bf 79.3} & {\bf 0.152} \\ \hline
\end{tabular}
\label{tab:results}
\end{center}
\end{table}
\subsection{Discussion on Results}
\label{sec:discussion_results}
In this section, we compare GeThR-Net with various baseline methods (Section \ref{sec:methods_compared}) and several recent state-of-the-art methods on three different datasets. The results corresponding to all the baselines and the proposed approach are summarized in Table~\ref{tab:results}. In the first two slabs of the table, results from individual modalities are shown using the temporal and non-temporal components. In the next three slabs, results for different fusion strategies across modalities are shown for both the temporal and non-temporal components. In the final slab of the table, results obtained from the proposed temporally hybrid component and GeThR-Net are shown. 
\noindent
\begin{itemize}[noitemsep,leftmargin=*]

\item{\bf UCF-101~\cite{soomro2012ucf101}:} For UCF-101, we report the test video classification accuracy. GeThR-Net achieves an absolute improvement of 3.1\%, 2.7\% and 4.6\% over Temporal-AM (late fusion), TemporalEtoE-AM (late fusion) and Temporal-AM (early fusion) baselines respectively. This empirically shows that the proposed approach is significantly better in capturing the complementary temporal aspects of different modalities compared to the late and early fusion based methods. GeThR-Net also gives an absolute improvement of 0.9\% over a strong baseline method of combining temporal and non-temporal aspects of different modalities (Temporal-AM+Non-Temporal-AM). This further establishes the efficacy of the proposed architecture. We also compare the results produced by GeThR-Net with several recent papers which reported results on UCF-101 (see Table \ref{tab:results_ucf-101}). Out of the seven approaches we compare, we are better than five of them and comparable to two (\cite{WangXW015} and \cite{wu2015modeling}) of them. As pointed out earlier, the goal of this paper is to develop a general deep learning framework which can be used for multimodal fusion in different kinds of tasks. The results on UCF-101 clearly shows that GeThR-Net can be effectively used for the short action recognition task (average duration 6-7 seconds). 

\item{\bf CCV~\cite{escalera2013multi}:} We also perform experiments on the CCV dataset to show that GeThR-Net can also be used for longer action recognition (average duration 80 seconds). In this dataset, we report the mean average precision (in a scale of 0-100) for all the algorithms which we compare. In CCV also, GeThR-Net is better than Temporal-AM (late fusion), TemporalEtoE-AM (late fusion) and Temporal-AM (early fusion) baselines by an absolute mAP of 4.3, 6.8 and 6.2 respectively. However, GeThR-Net performs comparable (mAP of 79.3 compared to 79.2) to a strong baseline method of combining temporal and non-temporal aspects of different modalities (Temporal-AM+Non-Temporal-AM). We also wanted to compare GeThR-Net with several recent approaches which also reported results on the CCV dataset. However, a fair comparison was not possible because several videos from CCV were unavailable from youtube. We used only 7,751 videos for training and testing as opposed to 9,317 videos in the original dataset. In spite of that, to get an approximate idea about how GeThR-Net performs compared to these methods, we provide some comparisons. The mAP reported on CCV by some of the recent methods are: 70.6~\cite{Wu2014}, 64.0~\cite{ye2012robust}, 63.4~\cite{Ma2014}, 60.3~\cite{Xu2013}, 68.2~\cite{Liu2013}, 64.0~\cite{MVA:audiovisual} and 83.5~\cite{wu2015modeling}. We perform better (mAP of 79.3) than six of these methods. 

\item{\bf MMG~\cite{escalera2013multi}:} In this dataset, we report the normalized edit distance (lower is better)~\cite{escalera2013multi} corresponding to each method. The normalized edit distance obtained by GeThR-Net is lower than the other multimodal baselines such as Temporal-AM (late fusion), TemporalEtoE-AM (early fusion), Temporal-AM (late fusion) and Temporal-AM+NonTemporal-AM by 0.004, 0.003, 0.038  and 0.003 respectively. We are also significantly better than modality specific temporal baselines, e.g.: GeThR-Net gives a normalized edit distance of only 0.152 compared to 0.284 and 0.361 produced by Temporal-M1 (audio) and Temporal-M2 (skeleton) respectively. The results on this dataset demonstrates that GeThR-Net performs well in fusing multimodal information from audio-MFCC and skeleton. The edit distance obtained from GeThR-Net is one of the top-three edits distances reported in the Chalearn-2013 multimodal gesture recognition competition~\cite{escalera2013multi}.  
\end{itemize}
\begin{table}[t]
\begin{center}
\caption{Comparison of GeThR-Net with state-of-the-art methods on UCF-101.}
\begin{tabular}{|p{0.10\textwidth}|p{0.125\textwidth}|p{0.125\textwidth}|p{0.11\textwidth}|p{0.10\textwidth}|p{0.125\textwidth}|p{0.125\textwidth}|p{0.12\textwidth}|}
\hline
 IDT + FV~\cite{Wang_2013_ICCV} & IDT + HSV\cite{PengWWQ14} & Two-stream~\cite{SimonyanZ14} & LSTM~\cite{NgHVVMT15} & TDD + FV~\cite{WangQT15a} & Two-stream2~\cite{WangXW015} & Fusion~\cite{wu2015modeling} & GeThR-Net \\ \hline \hline
85.9 & 87.9 & 88.0 & 88.6 & 90.3 & 91.4 & 91.3 & 91.1 \\ \hline
\end{tabular}
\label{tab:results_ucf-101}
\end{center}
\end{table}
From the results on these datasets, it is clear that GeThr-Net is effective in fusing different kinds of multimodal information and also applicable to different end-tasks such as short action recognition, long action recognition and gesture recognition. That empirically shows the generalizability of the proposed deep network.
\section{Conclusion}
In this paper, we propose a novel deep neural network called GeThR-Net for multimodal temporal information fusion. GeThR-Net has a temporally hybrid recurrent neural network component that models modality specific temporal dynamics as well as the temporal dynamics in a multimodal feature space. The other components in the GeThR-Net are used to capture the non-temporal information. We perform experiments on three different action and gesture recognition datasets and show that GeThR-Net performs well for any general multimodal fusion task. The experimental results are performed on four different modalities with maximum three modality fusion at a time. However, GeThR-Net can be used for any kind of modality fusion without any upper bound on the number of modalities that can be combined.

\clearpage

\bibliographystyle{splncs03}
\bibliography{sig-alternate-sample}
\end{document}